\documentclass[journal]{IEEEtran}
\usepackage{dsfont}

\ifCLASSINFOpdf
  \usepackage[pdftex]{graphicx}
  \graphicspath{{./figures/}}
\else
\fi

\ifCLASSINFOpdf
   \usepackage[pdftex]{graphicx}
   \graphicspath{{./figures}}
\else
\fi
\usepackage{amsmath}
\usepackage{algorithm}
\usepackage{algpseudocode}
\usepackage{array}
\usepackage{url}
\usepackage{tikz}

\newcommand{\Fo}{\mathbf{F}}

\newcommand{\x}{{\mathbf x}} 
 
\newcommand{\y}{{\mathbf y}} 
\newcommand{\A}{{\mathbf A}}

\newcommand{\eg}{\textit{e.g.}\ }
\newcommand{\ie}{\textit{i.e.}\ }

\usepackage{xcolor}
\definecolor{p_green}{HTML}{33CC66}
\definecolor{p_gold}{HTML}{CE9430}
\definecolor{p_blue}{HTML}{5083BF}

\begin{document}

\title{Memory-efficient Learning \\for Large-scale Computational Imaging}

\author{Michael~Kellman,
        Kevin~Zhang,
        Jon~Tamir,
        Emrah~Bostan,
        Michael~Lustig,
        and~Laura~Waller}

\markboth{Journal of \LaTeX\ Class Files,~Vol.~14, No.~8, August~2015}%
{Shell \MakeLowercase{\textit{et al.}}: Bare Demo of IEEEtran.cls for IEEE Journals}

\maketitle

\begin{abstract}
Critical aspects of computational imaging systems, such as experimental design and image priors, can be optimized through deep networks formed by the unrolled iterations of classical model-based reconstructions (termed physics-based networks). However, for real-world large-scale inverse problems, computing gradients via backpropagation is infeasible due to memory limitations of graphics processing units. In this work, we propose a memory-efficient learning procedure that exploits the reversibility of the network's layers to enable data-driven design for large-scale computational imaging systems. We demonstrate our method on a small-scale compressed sensing example, as well as two large-scale real-world systems: multi-channel magnetic resonance imaging and super-resolution optical microscopy.
\end{abstract}

% \begin{IEEEkeywords}
% \end{IEEEkeywords}

% \fi
\IEEEpeerreviewmaketitle

\section{Introduction}
\label{sec:intro}

Computational imaging systems (\eg tomographic systems, computational optics, magnetic resonance imaging (MRI)) jointly design software and hardware to retrieve information which is not traditionally accessible. Generally, such systems are characterized by how the information is encoded (forward process) and decoded (inverse problem) from the measurements. The decoding process is typically iterative in nature, alternating between enforcing data consistency and image prior knowledge. Recent work has demonstrated the ability to optimize computational imaging systems by unrolling the iterative decoding process to form a differentiable Physics-based Network (PbN)~\cite{Gregor:2010, sun2016deep, kellman2019physics} and then relying on a training dataset to learn the system's design parameters. Specifically, PbNs are constructed from the operations of image reconstruction algorithms (\eg proximal gradient descent or half quadratic splitting), where the iterations of the optimizer form the layers of the network. By including known structures and quantities, such as the forward model, data consistency, and signal prior, PbNs can be efficiently parameterized by only a limited number of learnable variables, thereby enabling an efficient use of training data~\cite{aggarwal2018modl} while still retaining the robustness and interpretability associated with conventional physics-based inverse problems. Commonly, standard signal prior models (\eg total variation) or the function that enforces consistency (\ie proximal operators) have been replaced by a learnable convolutional neural network~\cite{sun2016deep, aggarwal2018modl, diamond2017unrolled}. In addition to optimizing the image reconstruction algorithm using PbNs, one can also learn the data capture scheme (\ie  experimental design) by making the system parameters that form the measurements learnable~\cite{kellman2019physics, kellman2019data, Sitzmann:2018bd}).

Computational imaging systems present unique challenges for PbN implementation, due to the large size and dimensionality of variables that are decoded from the measurements. Training such a PbN relies on gradient-based updates computed using backpropagation (an implementation of reverse-mode differentiation~\cite{Griewank2018evalder}) for learning. As the quantity of decoded information grows, the memory required to perform backpropagation (via automatic differentiation) may exceed the memory capacity of the graphics processing unit (GPU).

Methods to save memory during backpropagation (forward recalculation, forward checkpointing, and reverse recalculation) trade off storage and computational complexity (\textit{i.e.} the amount of memory and time required for each unrolled layer)~\cite{Griewank2018evalder}. Rather than storing the whole computational graph required for auto-differentiation in memory, these methods reform the graph on an on-demand basis. For a PbN with $N$ layers, standard backpropagation stores the whole graph, achieving $\mathcal{O}(N)$ computational and storage complexity. Forward recalculation instead reforms unstored parts of the graph by reevaluating the operations of the network forward from the beginning. This achieves $\mathcal{O}(1)$ storage complexity, but has $\mathcal{O}(N^2)$ computational complexity because layers of the graph are recomputed from the beginning of the network, while backpropagation requires access to the layers in reverse order. Forward checkpointing saves variables every $K$ layers and forward-recalculates unstored layers of the graph from the closest stored variables (checkpoints), thus directly trading off computational, $\mathcal{O}(NK)$, and storage, $\mathcal{O}(N/K)$, complexity.

Reverse recalculation provides a practical solution to beat the trade-off between storage vs. computational complexities by reforming unstored layers of the graph in reverse order from the output of the network (in the same order as required for backpropagation), yielding $\mathcal{O}(N)$ computational and $\mathcal{O}(1)$ storage complexities. Recently, several reversibility schemes have been proposed for residual neural networks~\cite{behrmann2018invertible}, learning ordinary differential equations~\cite{chen2018neural}, and other specialized network architectures~\cite{gomez2017reversible, chang2018reversible}, including a PbN for MR images~\cite{putzky2019invert}.

Here, we propose a memory-efficient PbN learning procedure based on the concept of reverse recalculation and invertibility, enabling learning for large-scale applications in computational imaging. We describe how to compute gradients for learning using our method for PbNs formed from two representative optimization methods: proximal gradient descent and half quadratic splitting. Specifically, we detail how our memory-efficient learning for any PbN composed of gradient, proximal, and least-squares layers can be performed accurately and efficiently in time and memory. We highlight practical restrictions on the layers (\ie bijectivity) and present a hybrid scheme that combines our reverse recalculation methods with checkpointing to mitigate error accumulation due to numerical precision. We demonstrate our method by learning the design for a small-scale compressed sensing problem as a simple example, then show applications for two large-scale computational imaging systems: super-resolution optical microscopy (Fourier Ptychography) and 3D multi-channel MRI. In each of these applications, we are able to learn an optimized computational imaging system at a scale that was not previously possible. 
%This work is an extension of our recent conference abstract~\cite{kellman2019memory} and expands on its methodology and results.

\section{Background}
\label{sec:background}

\begin{figure}[t]
    \centering
    \includegraphics[width=8.89cm]{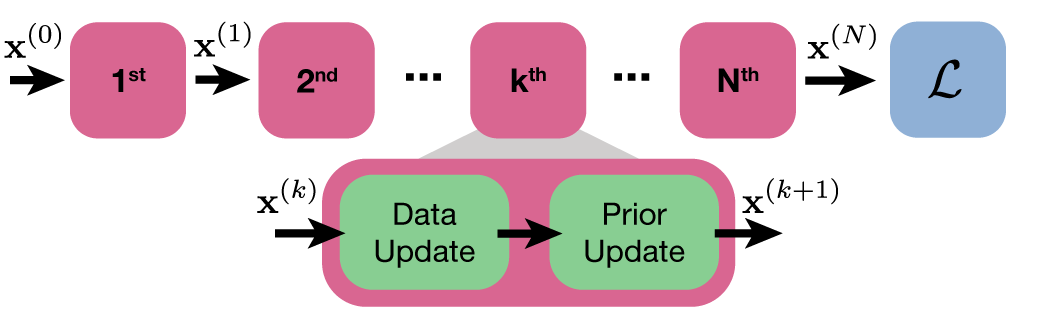}
    \caption{Physics-based Networks (PbNs) are formed by unrolling the iterations of an image reconstruction optimization. Each layer contains one iteration, made up of a data consistency update and signal prior update. The PbN input is the reconstruction's initialization, $\x^{(0)}$, and the output is the reconstructed image from the $N^{\text{th}}$ layer, which is fed into the learning loss, $\mathcal{L}$.}
    \label{fig:unrolled_network}
\end{figure}

The forward process for a typical computational imaging system describes how information about the image to be reconstructed, $\mathbf{x}$, is encoded into the measurements, $\mathbf{y}$. Specifically,
\begin{align}
    \mathbf{y} = \mathcal{A}(\mathbf{x}) + \mathbf{n},
    \label{eq:forward}
\end{align}
where $\mathcal{A}$ is the forward model that characterizes the measurement system physics and $\mathbf{n}$ is noise. The forward model is a continuous process, but is often represented by a discrete approximation. The inverse problem (\ie decoding) is commonly formulated as an optimization problem,
\begin{align}
    \mathbf{x}^{\star} = \arg\underset{\mathbf{x}}{\min}\ \mathcal{D}(\mathbf{x};\mathbf{y}) + \mathcal{P}(\mathbf{x}),
    \label{eq:decoder}
\end{align}
where $\mathcal{D}(\cdot)$ is a data consistency penalty and $\mathcal{P}(\cdot)$ is a signal prior penalty. When the noise, $\mathbf{n}$, is governed by a known noise model, the data consistency penalty can be written as the negative log-likelihood of the appropriate distribution. Proximal gradient descent (PGD) and half quadratic splitting (HQS) are two choices of algorithm for minimizing the objective in Eq.~\ref{eq:decoder} and can be used to form PbNs (Fig.~\ref{fig:unrolled_network}) that alternate between minimizing the data consistency and signal prior penalties. 

PGD is efficient in the case when $\mathcal{A}$ is non-linear and/or $\mathcal{P}(\x)$ is not smooth in $\x$ (\eg $\ell_1$, total variation). The PGD algorithm is composed of the following alternating steps:

\begin{align}
    \mathbf{z}^{(k)} &= \mathbf{x}^{(k)} - \alpha \nabla_\mathbf{x}\mathcal{D}(\mathbf{x}^{(k)};\mathbf{y}), \label{eq:pgd_grad} \\
    \mathbf{x}^{(k+1)} &= \texttt{prox}_{\mathcal{P}}(\mathbf{z}^{(k)}), \label{eq:pgd_prox}
\end{align}

\noindent where $\alpha$ is the gradient step size, $\nabla_{\mathbf x}$ is the gradient operator, $\text{prox}_{\mathcal{P}}$ is a proximal function that enforces the prior~\cite{parikh2014proximal}, and $\mathbf{x}^{(k)}$ and $\mathbf{z}^{(k)}$ are intermediate variables for the $k^\mathrm{th}$ iteration.

HQS is a more efficient algorithm when the forward model, $\mathcal{A}$, is linear and $\mathcal{P}(\x)$ is not smooth in $\x$. While similar to PGD in alternating between data consistency and prior updates, HQS instead performs a full model inversion rather than a single gradient step for the data consistency update:

\begin{align}
    \mathbf{z}^{(k)} &= \arg\underset{\mathbf{z}}{\min} \ \mathcal{D}(\mathbf{z};\mathbf{y}) + \mu \|\mathbf{z} - \mathbf{x}^{(k)}\|_2^2 \label{eq:hqs_dc} \\
    \mathbf{x}^{(k+1)} &= \texttt{prox}_{\mathcal{P}}(\mathbf{z}^{(k)}), \label{eq:hqs_prox}
\end{align}

\noindent where $\mu$ is a penalty parameter that weights the data consistency and signal prior penalties. When the noise has a normal distribution, Eq.~\ref{eq:hqs_dc} can be efficiently solved via the conjugate gradient (CG) method.

The structure of the PbN is determined by unrolling $N$ iterations (Fig.~\ref{fig:unrolled_network}) of the optimizer (Eq.~\ref{eq:decoder}) to form $N$ layers of a network (\eg for PGD, Eq.~\ref{eq:pgd_grad} and Eq.~\ref{eq:pgd_prox} form a single layer and for HQS, Eq.~\ref{eq:hqs_dc} and Eq.~\ref{eq:hqs_prox} form a single layer). The input to the network is the initial guess for the reconstructed image, $\mathbf{x}^{(0)}$, and the output is the resultant, $\mathbf{x}^{(N)}$. Commonly, the learnable parameters are optimized using gradient-based methods (\eg stochastic gradient descent or ADAM~\cite{kingma2014adam}) and machine learning toolboxes' (\eg PyTorch~\cite{paszke2017automatic}, Tensor Flow~\cite{tensorflow2015-whitepaper}) auto-differentiation functionalities are used to compute the gradients. Auto-differentiation creates a computational graph composed of the PbN's operations and stores intermediate variables in memory on the forward pass of the network. On the backward pass, auto-differentiation traces through the graph from the output to the input, computing the Jacobian-vector product for each operation.

\section{Methods}
\label{sec:methods}

\begin{figure}[t]
    \centering
    \includegraphics[width=8.89cm]{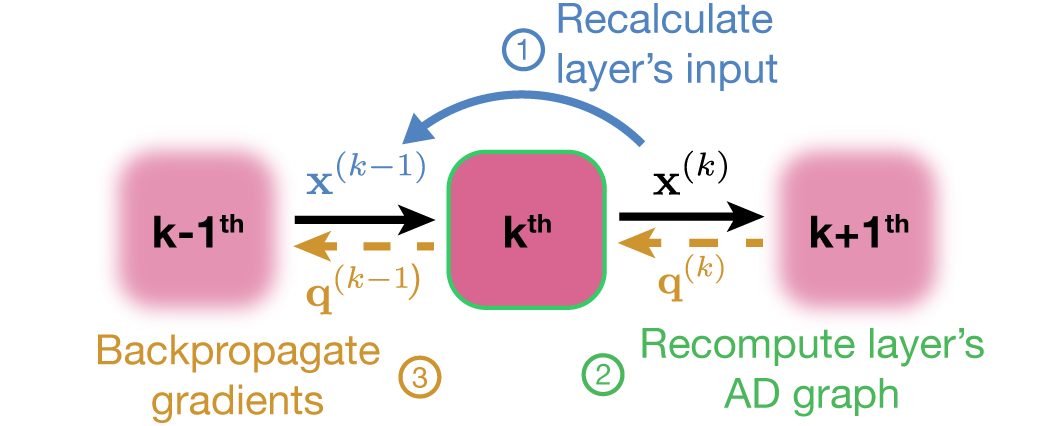}
    \caption{Memory-efficient learning procedure for a single layer: (1) recalculate the layer's input, $\mathbf{x}^{(k-1)}$, from the output, $\mathbf{x}^{(k)}$, by applying that layer's inverse operations. (2) Recompute the auto-differentiation graph for that single layer. (3) Backpropagate gradients, $\mathbf{q}^{(k)} = \partial \mathcal{L} / \partial  \mathbf{x}^{(k)}$, through the layer's auto-differentiation graph.}
    \label{fig:inverse_network}
\end{figure}

Our main contribution is to improve the storage and computational complexity of backpropagation for PbNs by treating the single large graph for auto-differentiation as a series of smaller graphs that can be reformed from the output in reverse order and are only required to be stored in memory one at a time. Consider a PbN, $\mathcal{F}$, composed of a sequence of layers, 

\begin{align}
    \mathbf{x}^{(k+1)} = \mathcal{F}^{(k)}\left(\mathbf{x}^{(k)}; \theta^{(k)}\right),
\end{align}

\noindent where $\mathbf{x}^{(k)}$ and $\mathbf{x}^{(k+1)}$ are the $k^{\text{th}}$ layer input and output, respectively, and $\theta^{(k)}$ are its learnable parameters. When performing reverse-mode differentiation, our method treats a PbN of $N$ layers as $N$ separate smaller graphs, generated on demand, processed and stored one at a time, rather than as a single large graph, thereby saving a factor $N$ in memory. As outlined in Alg.~\ref{alg:meld} and Fig.~\ref{fig:inverse_network}, we first recalculate the current layer's input, $\mathbf{x}^{(k-1)}$, from its output, $\mathbf{x}^{(k)}$, using $\mathcal{F}^{(k-1)}_{\text{inverse}}$ (Alg.~\ref{alg:meld} line~\ref{subalg:reverse}), and then form one of the smaller graphs by recomputing the output of the layer, $\mathbf{v}^{(k)}$, from the recalculated input (Alg.~\ref{alg:meld} line~\ref{subalg:forward}). To compute gradients, we then rely on auto-differentiation of each layer's smaller graph to compute the gradient of the loss, $\mathcal{L}$, with respect to $\mathbf{x}^{(k)}$ (denoted $\mathbf{q}^{(k)}$) (Alg.~\ref{alg:meld} line~\ref{subalg:backprop}) and $\nabla_{\theta^{(k)}} \mathcal{L}$ (Alg.~\ref{alg:meld} line~\ref{subalg:gradient}). The procedure is repeated for all $N$ layers in reverse order.
% As outlined in Alg.~\ref{alg:meld}, our method reverse recalculates the previous layer input from the current, constructs a graph for auto-differentiation, and then backpropagates the derivative of the loss wrt. $\mathbf{x}^{(k)}$ (denoted $\mathbf{q}^{(k)}$).
\begin{algorithm}
    \caption{Memory-efficient learning for physics-based networks}
    \label{alg:meld}
    \begin{algorithmic}[1]
        \Procedure{Memory-efficient Backpropagation}{$\mathbf{x}^{(N)}, \mathbf{q}^{(N)}$}
        \State $k \gets N$
        \For{$k > 0$}
        \State $\mathbf{x}^{(k-1)} \gets \mathcal{F}^{(k-1)}_{\text{inverse}}(\mathbf{x}^{(k)}; \theta^{(k-1)})$ \label{subalg:reverse}
        \State $\mathbf{v}^{(k)} \gets \mathcal{F}^{(k-1)}(\mathbf{x}^{(k-1)}; \theta^{(k-1)})$ \label{subalg:forward}
        \State $\mathbf{q}^{(k-1)} \gets \frac{\partial \mathbf{v}^{(k)}}{\partial \mathbf{x}^{(k-1)}}\mathbf{q}^{(k)}$ \label{subalg:backprop}
        \State $\nabla_{\theta^{(k)}} \mathcal{L} \gets \frac{\partial \mathbf{v}^{(k)}}{\partial \theta^{(k)}}\mathbf{q}^{(k)}$ \label{subalg:gradient}
        \State $k \gets k - 1$
        \EndFor
        \State \textbf{return} $\{\nabla_{\theta^{(k)}} \mathcal{L}\}_{k=0}^{N-1}$
        \EndProcedure
    \end{algorithmic}
\end{algorithm}

In order to perform the reverse-mode differentiation efficiently, our method must be able to compute each layer's inverse operation, $\mathcal{F}_{\text{inverse}}^{(k-1)}$. The remainder of this section overviews the procedures to invert gradient, proximal, and least-squares update layers.

\subsection{Inverse of gradient update layer}
\label{ssec:inverse_grad}

A common interpretation of gradient descent is as a forward Euler discretization of the continuous-time ordinary differential process~\cite{parikh2014proximal} gradient flow. Thus, the inverse of the gradient step layer (Eq.~\ref{eq:pgd_grad}) can be viewed as a backward Euler step,
\begin{align}
    \mathbf{x}^{(k)} &= \mathbf{z}^{(k)} + \alpha \nabla_\mathbf{x}\mathcal{D}(\mathbf{x}^{(k)};\mathbf{y}).
\end{align}
This implicit equation can be solved iteratively via the backward Euler method using a fixed point algorithm (Alg.~\ref{alg:grad_fp})~\cite{parikh2014proximal}. Convergence is guaranteed if
\begin{align}
    \text{Lip}\left(\alpha \nabla_\mathbf{x}\mathcal{D}(\mathbf{x};\mathbf{y})\right) &< 1, \label{eq:lipschitz}
\end{align}
where $\text{Lip}(\cdot)$ computes the Lipschitz constant of its argument~\cite{banach1922operations}. In the setting when $\mathcal{D}(\mathbf{x};\mathbf{y}) = \|\mathbf{A}\mathbf{x} - \mathbf{y}\|^2$ and the forward model, $\mathbf{A}$, is linear, this can be ensured if $\alpha < \frac{1}{\sigma_{max}(\mathbf{A}^H\mathbf{A})}$, where $\sigma_{max}(\cdot)$ computes the largest singular value of its argument. Finally, as given by Banach Fixed Point Theorem, the fixed point algorithm (Alg.~\ref{alg:grad_fp}) will have an exponential rate of convergence~\cite{banach1922operations}.
\begin{algorithm}
    \caption{Inverse for gradient layer}\label{alg:grad_fp}
    \begin{algorithmic}[1]
        \Procedure{Fixed Point Method}{$\mathbf{z},T$}
        \State $\mathbf{x} \gets \mathbf{z}$
        \For{$t < T$}
        \State $\mathbf{x} \gets \mathbf{z} + \alpha \nabla_\mathbf{x}\mathcal{D}(\mathbf{x};\mathbf{y})$
        \State $t \gets t + 1$
        \EndFor
        \State \textbf{return} $\mathbf{x}$
        \EndProcedure
    \end{algorithmic}
\end{algorithm}

\subsection{Inverse of proximal update layer}
\label{ssec:inverse_prox}
The proximal update (Eq.~\ref{eq:pgd_prox} and Eq.~\ref{eq:hqs_prox}) is defined by the following optimization problem~\cite{parikh2014proximal}:
\begin{align}
    \text{prox}_{\mathcal{P}}(\mathbf{z}^{(k)}) = \arg\underset{\mathbf{v}}{\min}\ \frac{1}{2}\|\mathbf{v} - \mathbf{z}^{(k)}\|_2^2 + \mathcal{P}(\mathbf{v}).
    \label{eq:prox}
\end{align}

For differentiable $\mathcal{P}(\cdot)$, the solution to Eq.~\ref{eq:prox} gives,

\begin{align}
    \mathbf{x}^{(k+1)} = \mathbf{z}^{(k)} - \nabla_\mathbf{x} \mathcal{P}(\mathbf{x}^{(k+1)}).
\end{align}
In contrast to the gradient update layer, the proximal update layer can be thought of as a backward Euler step~\cite{parikh2014proximal}. This allows its inverse to be expressed as a forward Euler step,
\begin{align}
     \mathbf{z}^{(k)} = \mathbf{x}^{(k+1)} + \nabla_\mathbf{x} \mathcal{P}(\mathbf{x}^{(k+1)}),
\end{align}
when the proximal function is bijective (\eg $\text{prox}_{\ell_2}$). If the proximal function is not bijective (\eg $\text{prox}_{\ell_1}$), the inversion is not straightforward; however, in many cases we can substitute it with a bijective function with similar behavior. For example, soft thresholding, the proximal operator of $\ell_1$ norm, is not bijective, but can be made so by adding a small slope.

\subsection{Inverse of least squares update layer}
\label{ssec:linsys}

For HQS, the minimization in Eq.~\ref{eq:hqs_dc} performs the data consistency update. When Eq.~\ref{eq:forward} is linear, the solution to this minimization is:

\begin{align}
    \mathbf{z}^{(k+1)} = \left( \mathbf{A}^H\mathbf{A} + \mu \mathbf{I} \right)^{-1} (\mathbf{A}^H\mathbf{y} + \mu \mathbf{x}^{(k)}).
    \label{eq:rlsq}
\end{align}

When $\mathbf{A}$ models a linear translation invariant system, it is a circular convolution and Eq.~\ref{eq:rlsq} can be computed in closed form via frequency division. However, often computational imaging systems represent $\mathbf{A}$ not as an explicit matrix but as a series of operators. In this case, the inversion can be efficiently computed using a conjugate gradient (CG) method.

The inverse of this layer is found in closed form:

\begin{align}
    \mathbf{x}^{(k)} = \frac{1}{\mu} \left(\left( \mathbf{A}^H\mathbf{A} + \mu \mathbf{I} \right) \mathbf{z}^{(k+1)} - \mathbf{A}^H\mathbf{y}\right).
\end{align}

\noindent When using a CG method to solve Eq.~\ref{eq:rlsq}, the inverse is accurate only if CG performs the model inversion accurately. This is a possible source of numerical error that is further discussed in Sec.~\ref{sec:hybrid}.

\begin{figure}[t]
    \centering
    \includegraphics[width=8.89cm]{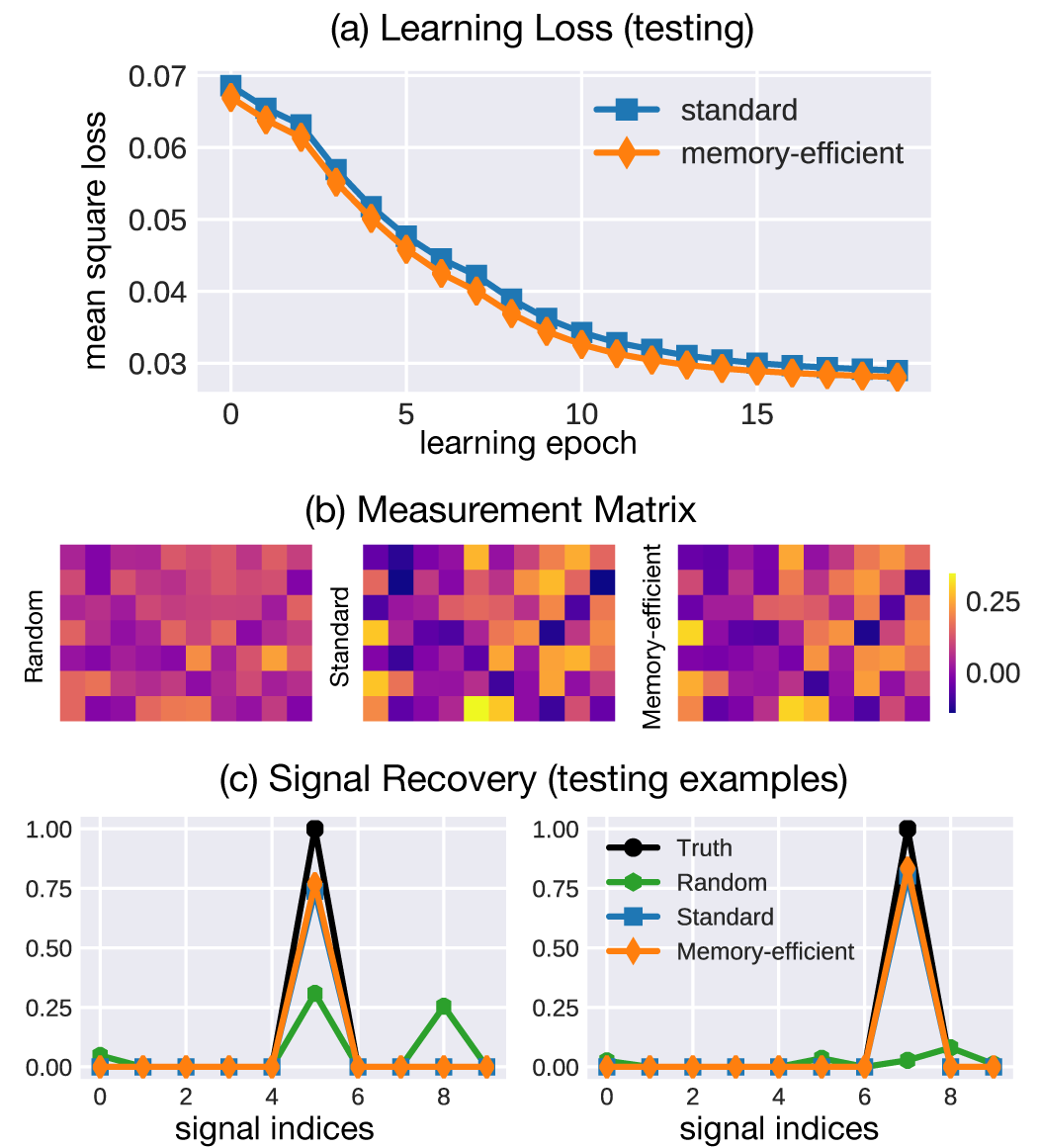}
    \caption{Learned measurements for 1D compressed sensing: (a) Mean testing loss for learning using standard and memory-efficient learning techniques. (b) Initial (Gaussian randomly distributed) and learned measurement matrices using standard and memory-efficient techniques. (c) Two testing examples of reconstructions with random and learned measurement schemes, demonstrating both improved signal recovery using the learned measurements in comparison to the random measurements and similarity between standard and memory-efficient learning (while requiring $4.1$KB, $\sim 800\times$ less memory than standard backpropagation).}
    \label{fig:cs_example}
\end{figure}

\begin{figure}[t]
    \centering
    \includegraphics[width=8.89cm]{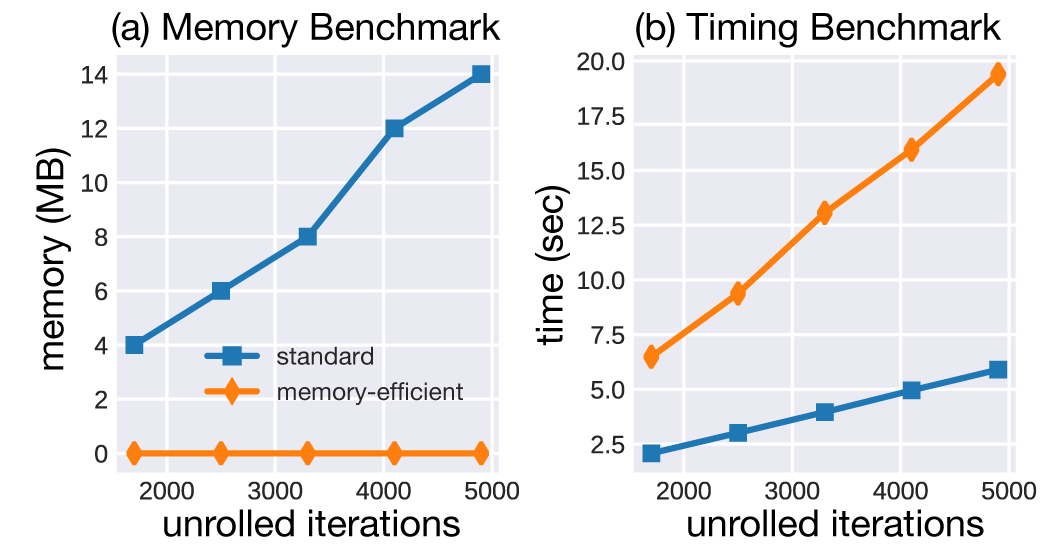}
    \caption{Performance benchmarks for compressed sensing example: (a) Memory required vs number of unrolled iterations in the PbN, for standard and memory-efficient learning. (b) Time required vs number of unrolled iterations in the PbN for standard and memory-efficient learning. Memory-efficient learning has a constant memory requirement, while standard backpropagation requires linearly increasing amounts of memory as the number of unrolled iterations increases. This improvement in memory requirements comes at a cost of somewhat longer compute times.}
    \label{fig:cs_benchmark}
\end{figure}

\section{Hybrid Reverse Recalculation and Checkpointing}
\label{sec:hybrid}

Reverse recalculation of the unstored variables is non-exact, as the operations to calculate the variables are not identical to forward calculation. The result is numerical error between the original forward and reverse calculated variables. As more iterations are unrolled, numerical errors can accumulate.

To mitigate these effects, we can use checkpointing. Some of the intermediate variables can be stored from forward calculation and used in substitution for the recalculated variables, that could incur accumulated numerical errors. Memory permitting, as many checkpoints as possible should be stored to ensure accuracy while performing reverse recalculation. Due to the size of the intermediate variables, large-scale PbNs cannot afford to store all variables required for reverse-mode differentiation, but it is often possible to store a few as checkpoints.

Further, when enough iterations of the reconstruction optimization (Eq.~\ref{eq:decoder}) are unrolled, convergence of the intermediate variables can often be observed. When this occurs, inversion of each layer's operations (Alg.~\ref{alg:meld} line~\ref{subalg:reverse}) becomes ill-posed. For example, when PGD converges the gradient of the reconstruction loss will be zero, thus Alg.~\ref{alg:grad_fp} will return its input and the inversion will fail.

Checkpointing can again be used to reduce these effects. If convergence behavior is observed, then checkpoints can be stored during later layers to correct inversion error. Economically, checkpoints should be placed closer together for later layers and less frequently for earlier layers (this further discussed in Sec.~\ref{sec:discussion}).
% From a practical perspective, we do not have a rule to inform when to place checkpoints, however, if the difference between layer's falls below a threshold, then storage 
%  While more checkpoints might be required in these settings, they do not have to equally spaced.

\section{Results}
\label{sec:results}

We first demonstrate our memory-efficient learning method with a small-scale compressed sensing system as an example, then with two real-world large-scale applications. In the compressed sensing example, we learn the measurement matrix to improve reconstruction performance and empirically test our method's storage and computational complexities. In the first of our large-scale applications, we improve the image quality for multi-channel accelerated MRI by learning better signal priors to regularize the reconstruction. In the second, we improve the temporal resolution of super-resolution microscopy (Fourier Ptychography) by learning the system's experimental design.

% Both examples can be demonstrated at smaller scales, thereby allowing us to benchmark the accuracy of our method against standard backpropagation, however, learning for these system parameters at realistic scales will require our memory-efficient learning method.

\begin{figure*}[t]
    \centering
    \includegraphics[width=18.19cm]{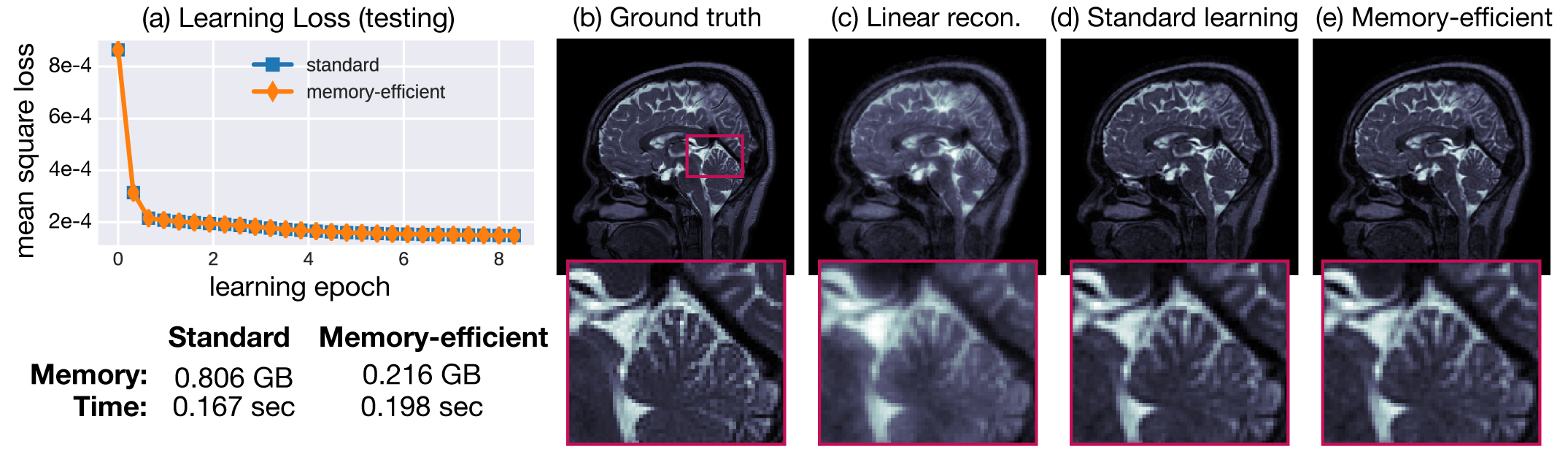}
    \caption{Learned priors for multi-channel 2D under-sampled MRI: (a) Mean testing loss is similar for both standard backpropagation and memory-efficient learning. (b) Ground truth reconstruction using fully sampled measurements, (c) linear parallel imaging reconstruction (no prior), (d) PbN reconstruction learned using standard backpropagation and (e) PbN reconstruction learned using memory-efficient learning (3.7$\times$ reduced memory requirement, 1.2$\times$ increase in compute time). Insets highlight fidelity of high-resolution features and noise reduction in both of the learned designs, as compared to the CG reconstruction. Reported memory and time required is for a single learning update with batch size one.}
    \label{fig:mri_learning}
\end{figure*}

\begin{figure*}[t]
    \centering
    \includegraphics[width=18.19cm]{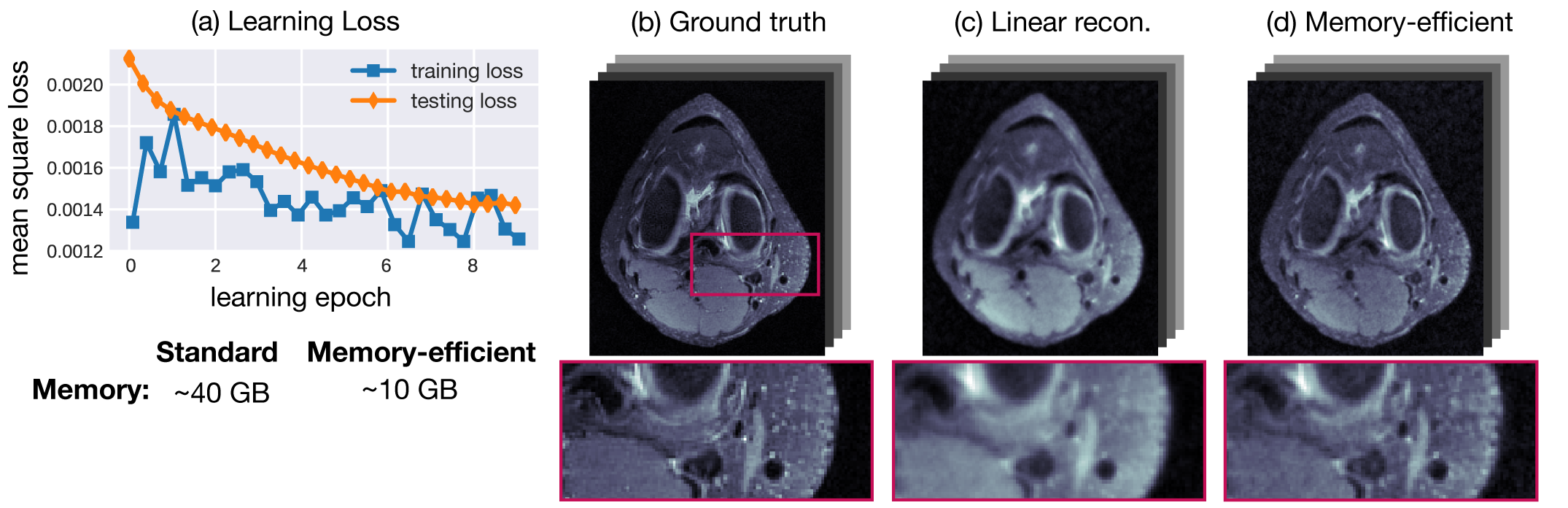}
    \caption{Learned priors for multi-channel under-sampled 3D MRI: (a) Mean training and testing loss for learning with our proposed memory-efficient technique. (b) One slice of ground truth 3D reconstruction using fully sampled measurements, (c) linear parallel imaging reconstruction (no prior), (d) PbN reconstruction using memory-efficient learning with $\sim10$GB of memory. Standard learning is not shown because it requires more memory than would fit on our GPU.}
    \label{fig:mri_ls_learning}
\end{figure*}

\subsection{Learned measurements for compressed sensing}
\label{ssec:cs_example}

Compressed sensing combines random measurements and regularized optimization to reduce the sampling requirements of a signal below the Nyquist rate~\cite{donoho2006}. It has seen practical success in many fields (\eg MRI~\cite{bib:lustig2007hl}, holography~\cite{brady2009compressive}, optical imaging~\cite{duarte2008single}). A natural question to ask is, which measurements provide the best signal recovery for a class of signals? Specifically, we recover arbitrary one-sparse signals from linear measurements and learn the linear measurement matrix with a PbN. We learn a set of 7 coded 1D masks; each scalar measurement is the dot product of a mask with the signal and optimize recovery of the signal in terms of mean square error. The dimensions and scale of compressed sensing problem we setup are small scale. This is intended to rapidly demonstrate the accuracy of our method and to measure the storage and computational complexity of standard and memory-efficient learning techniques. The PbN is constructed by unrolling PGD for the reconstruction loss,

\begin{align}
    \x^\star = \arg\underset{\x}{\min} \|\A\x - \y\|^2_2 + \lambda \|\x\|_1.
\end{align}

\noindent where $\A \in \mathds{R}^{7 \times 10}$, $\x \in \mathds{R}^{10}$ is a one-sparse signal, $\y \in \mathds{R}^{7}$ is a measurement signal, and $\lambda$ trades off the data consistency and sparsity prior penalties. The PbN is formed from $800$ unrolled iterations of PGD with a step size of $0.05$ and $\lambda=0.06$. For our method, a modified soft thresholding function is used as the proximal operator, where a small slope (on the order of $1e^{-6}$) is added to the zeroed region to make it an invertible function (as discussed in Sec.~\ref{ssec:inverse_prox}). Training was conducted for $20$ epochs with $20$ training data points, batch size of $4$, and learning rate of $1e^{-2}$ using ADAM~\cite{kingma2014adam}. $50$ checkpoints are used to mitigate error due to numerical precision (as discussed in Sec.~\ref{sec:hybrid}).

Figure~\ref{fig:cs_example} shows a comparison between the testing loss for standard and memory-efficient learning techniques, initial random and optimized measurement matrices, and several testing data points for the ground truth with the signal recovered using the learned and initial matrix (random Gaussian variable). As seen in Fig.~\ref{fig:cs_example}c, the learned measurement matrices have better signal recovery than the random matrix. The learned measurements and signal recovery using our method and standard learning are similar, but we use $\sim 800\times$ less memory. Where as our method uses a modified soft thresholding function, standard learning uses the ordinary version of the function. Learning results are comparable (Fig.~\ref{fig:cs_example}) between the uses of the two functions, suggesting the affect of the modification is negligible; further discussion is included in Sec.~\ref{sec:discussion}.

In Fig.~\ref{fig:cs_benchmark}, we empirically benchmark the storage and computational complexities of our method and standard backpropagation. As predicted, our method requires a constant amount of memory independent of the number of unrolled iterations, while standard learning requires memory linear in the number of layers. In terms of time, both methods exhibit linear time complexity, with our method being slower than standard backpropagation by a factor of $\sim2.5\times$.

\subsection{Learned priors for multi-channel MRI}
\label{ssec:mri}

As our first real-world example, we look at MRI, a powerful medical imaging modality that non-invasively captures rich biophysical information without ionizing radiation. Since MRI acquisition time is often directly proportional to the number of acquired measurements, reducing measurements leads to immediate impact on scan time, patient throughput, and enables capturing fast-changing physiological dynamics. Multi-channel MRI is the standard of care in clinical systems and uses multiple receive coils distributed around the body to acquire measurements in parallel, termed parallel imaging, it reduces the total number of required acquisition frames for decoding~\cite{bib:pruessmann1999ys}. Further, scan time and noise amplification, can be additionally reduced by relying on signal prior knowledge, allowing the undersampling the acquisition frames (\ie with compressed sensing~\cite{bib:lustig2007hl}). Recently, PbNs have been developed to learn the signal priors, achieving state-of-the-art performance for multi-channel accelerated MRI~\cite{Hammernik:2017ku, aggarwal2018modl}. However, the PbNs are limited in network size and number of unrolled iterations due to the amount of memory required for training. This is an especially prominent problem when moving to high-dimensional problems (\eg 3D anatomical imaging, temporal dynamics, etc.). Our memory-efficient learning reduces memory footprint at training time, thereby enabling learning for larger problems.

To validate our method, we first show results for the 2D problem in~\cite{aggarwal2018modl}, which has small enough memory requirements for the standard backpropagation to fit on our GPUs. The PbN is formed from $4$ unrolled iterations of the HQS method and replaces the proximal operator with a Resnet to enforce the signal prior~\cite{aggarwal2018modl, zhang2017learning}. The data consistency layer (Eq.~\ref{eq:hqs_dc}) is defined from $\mathcal{D}(\x;\y) = \|\mathbf{P}\mathbf{F}\mathbf{S}\x - \y\|^2_2$, where $\mathbf{S}$ are the multi-channel coil sensitivities, $\mathbf{F}$ denotes Fourier transform, and $\mathbf{P}$ is the undersampling mask used for compressed sensing. The proximal operator (Eq.~\ref{eq:hqs_prox}) is represented using a learnable invertible residual convolutional neural network (RCNN)~\cite{he2016deep,gomez2017reversible,behrmann2018invertible} composed of a 5-layer CNN where each layer has $64$ channels and filters of $3 \times 3$. The RCNN's learnable parameters are shared between each PbN layer.

We learn to reconstruct $256 \times 320$ slices with measurements from $8$ channels and variable density Poisson Disc Fourier undersampling at a rate of $4\times$. Data used for training and testing is from~\cite{aggarwal2018modl}, where ground truth brain images are used and data is synthetically generated given the sensitivity and undersampling masks. Training was conducted for $10$ epochs with $20$ training data sets, a batch size of $4$, and a learning rate of $1e^{-5}$ using ADAM~\cite{kingma2014adam}. In Fig.~\ref{fig:mri_learning}, we compare image reconstructions using the priors learned by standard and memory-efficient learning. As shown in Fig.~\ref{fig:mri_learning}a, testing losses and image reconstruction quality are similar for both methods (Fig.~\ref{fig:mri_learning}d,e). Our method uses $4.82\times$ less memory, while only requiring a $1.09\times$ increase in time.

Finally, we demonstrate our method's ability to learn priors for a 3D volume reconstruction from under-sampled multi-channel measurements - a problem that does not fit within typical GPU memory limits. Specifically, we reconstruct volumes of $50 \times 256 \times 320$ with measurements from $8$ channels and variable density Poisson Disc undersampling at a rate of $4\times$. Data used is from~\cite{aggarwal2018modl} and is augmented to create more training examples by cropping down larger volumes to $50 \times 256 \times 320$. We use a similar PbN architecture as before for the reconstruction and training parameters, but now with a RCNN with 3D filters ($3 \times 3 \times 3$) and $32$ channels. This model would ordinarily require $\sim 40$GB of memory using standard backpropagation ($\sim 10$GB per unrolled iteration), but only requires $\sim 10$GB of memory using our method. In Fig.~\ref{fig:mri_ls_learning}, we show results of the learning loss and a single slice of the reconstructed volumes from the ground truth (fully sampled), conjugate gradient (no learning or signal prior), and after learning priors with our memory-efficient learning scheme.

\begin{figure*}[t]
    \centering
    \includegraphics[width=18.19cm]{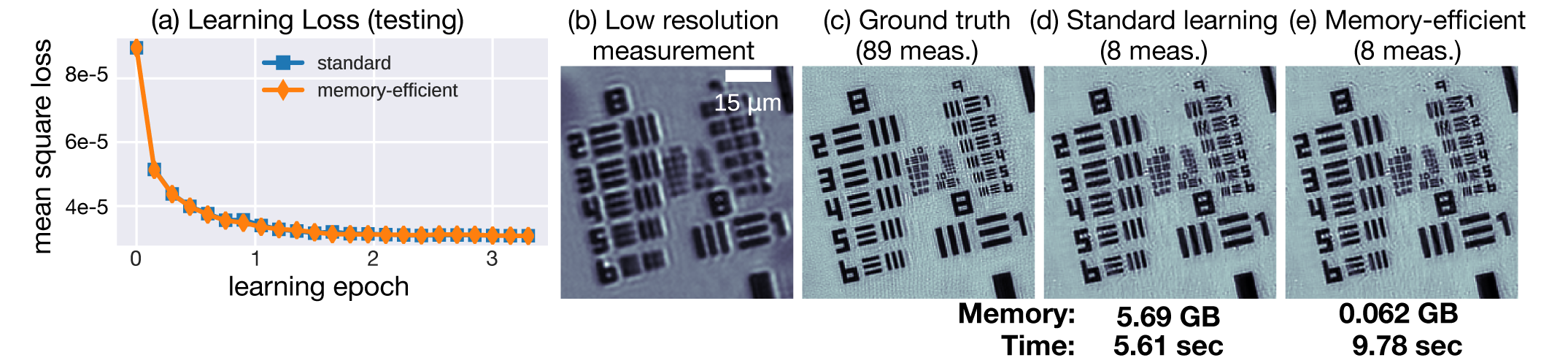}
    \caption{Learned illumination design for Fourier Ptychographic Microscopy (FPM): (a) Mean testing loss is similar for both standard backpropagation and memory-efficient learning. (b) Example low-resolution measurement, (c) ground truth reconstruction using all 89 LED measurements to perform $3.1\times$ super resolution, (d) reconstruction from only 8 measurements learned using standard backpropagation and (e) memory-efficient learning  (92$\times$ reduced memory requirement, 1.7$\times$ increase in compute time). Reported memory and time required is for a single learning update with batch size one.}
    \label{fig:ss_fpm_cmp}
\end{figure*}

\begin{figure*}[t]
    \centering
    \includegraphics[width=18.19cm]{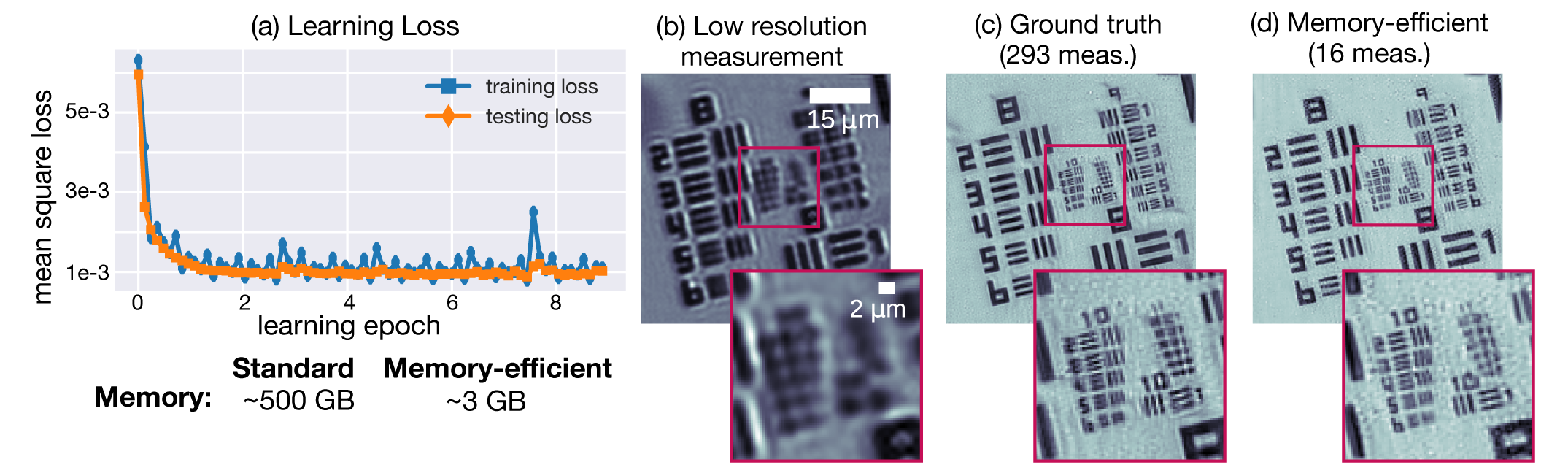}
    \caption{Large-scale learned illumination design for FPM: (a) Training and testing loss for memory-efficient learned design. (b) Example low-resolution measurement, (c) ground truth reconstruction using all 293 LED measurements to perform $4.2\times$ super resolution, (d) reconstruction from only 16 measurements learned using memory-efficient learning with $\sim3$GB memory. Standard learning is not shown because it would require $\sim500$GB of memory, which is not available on our GPU. Insets highlight high-resolution features.}
    \label{fig:ls_fpm}
\end{figure*}

\subsection{Learned experimental design for Fourier Ptychographic Microscopy}
\label{ssec:fpm}

Bright-field microscopy is a standard method for imaging biological samples \textit{in vitro}. As with most microscopes, one must trade-off field-of-view (FoV) and resolution. Fourier Ptychographic Microscopy (FPM)~\cite{Zheng:2013gq} is a super-resolution (SR) method that computationally reconstructs gigapixel-scale images with both large FoV and high resolution from a series of low-resolution images acquired with different illumination settings. The illumination patterns can be conveniently created by a programmable LED array source~\cite{phillips2017quasi}. The system's dependence on many measurements limits its ability to image live fast-moving biology, so multiplexing schemes for reducing the number of measurements have been proposed~\cite{Tian:2015er}. Recently, state-of-the-art performance was achieved by forming a PbN and learning its experimental design (the LED array patterns)~\cite{kellman2019data, kellman2019physics}. However, the PbN was limited in scale due to GPU memory constraints; terabyte-scale memory would be required for learning patterns with all of the LEDs. Here, we show that our proposed memory-efficient learning framework reduces the necessary memory to only a few gigabytes, thereby enabling full-scale learning on a consumer-grade GPU, which in turn allows us to achieve higher factors of super resolution.

The PbN for learning FPM LED source patterns is formed from the following phase retrieval optimization:

\begin{align}
    \x^{\star} = \arg \underset{\x}{\min} \sum_{k=1}^{K} \Bigl\|\y_{\mathrm{m}_k} - \sum_{l=1}^{L} c_{kl} | \A_l\x|^2 \Bigr\|_2^2 \text{,}
    \label{eq:fpm_recon}
\end{align}

\noindent where $\y_{\mathrm{m}_k}$ is the $k^{\text{th}}$ multi-LED measurement, $\A_l = \Fo^H\mathbf{P}_l\Fo$ is the forward model for the $l^{\text{th}}$ LED~\cite{kellman2019data, Zheng:2013gq}, $\mathbf{P}_l$ is the microscope's pupil function for the $l^{\text{th}}$ LED, $\Fo$ denotes 2D Fourier transform, and $c_{kl}$ is the learnable brightness for the $l^{\text{th}}$ LED in the $k^{\text{th}}$ measurement. A PbN is formed from $N$ unrolled iterations of gradient descent. We then  minimize the loss between the output of the PbN and the ground truth to learn LED brightnesses over the dataset.

We again start by validating our method's accuracy on a small-scale problem that fits in GPU memory using standard learning. We reproduce results in \cite{kellman2019data}, learning illumination patterns for eight measurements, which gives $3.1 \times$ resolution improvement and $10 \times$ faster data capture. We set $T=4$, the number of fixed point iterations to invert gradient layers, and checkpoints every $10$ unrolled iterations. The testing loss between our method and standard learning are similar (Fig.~\ref{fig:ss_fpm_cmp}a), and the SR reconstructions with learned designs using standard (Fig.~\ref{fig:ss_fpm_cmp}d) and memory-efficient (Fig.~\ref{fig:ss_fpm_cmp}e) methods are both similar to the `ground truth' reconstruction using 89 measurements (Fig.~\ref{fig:ss_fpm_cmp}c). Our memory-efficient learning approach, however, reduces memory required from $5.69$GB to $0.062$GB, with compute time increasing by less than a factor of $2 \times$. Hence, our method produces comparable quality results as the standard learning, but with significantly reduced (more than 91$\times$) memory requirements.

Next, we use our memory-efficient learning scheme to solve a larger-scale problem than was previously possible. For FPM, that means using all 293 LEDs to achieve a higher factor of super resolution ($4.2 \times$). $200$ iterations are unrolled to create the PbN, we set $T=4$, and checkpoints every $13$ unrolled iterations. For this problem, standard backpropagation would require $\sim 500$GB of memory, while our method only requires $\sim 3$GB (using 15 checkpoints). In Fig~\ref{fig:ls_fpm}, we demonstrate our learned design's ability to reduce the number of measurements required from $293$ to $16$, demonstrating $20\times$ faster data capture with comparable image quality to ground truth. 

\section{Discussion}
\label{sec:discussion}

Our proposed memory-efficient learning opens the door to using unrolled physics-based networks for learning in applications that are not otherwise possible due to GPU memory constraints, without a significant increase in training time. While we have specialized the procedure to the layers of PbNs formed from PGD and HQS methods, the update layers we describe form the fundamental building blocks of many larger PbNs (\eg unrolling the updates of alternating minimization). 

For each layer, sufficient conditions for invertibility must be met. This limitation is clear in the case of a gradient descent block with an evolving step size, as the Lipschitz constant may no longer satisfy Eq.~\ref{eq:lipschitz}. Furthermore, the convergent behavior of reconstruction optimization (Eq.~\ref{eq:decoder}) makes accurate reverse recalculation ill-posed and can cause numerical error accumulation (as outlined in Sec.~\ref{sec:hybrid}). This is not an issue for many PbNs as they are not deep enough to reach numerical convergence. In the case when convergent behavior is observed, checkpoints should be used, however, when to place checkpoints is not clear. A possible option is to measure the difference between successive intermediate variables on the forward pass of the network. If that quantity falls below a threshold, then the optimization is approaching convergence and checkpoints should be placed more often to mitigate the accumulation of error on the reverse pass.

% more often in the later layers of the network to ensure accuracy, while they can be less frequently for earlier layers (where inversion is more well posed).

In some situations the relationship between storage and computational complexity can be traded off with accuracy. For gradient descent layers, the fixed point method outlined in Alg.~\ref{alg:grad_fp} is used to invert and, if not run to convergence, the inversion will be less accurate. When the Lipschitz constant of the gradient operator is large, more iteration (a larger value of $T$) will be required to accurately invert the layer. Unfortunately, the ideal Lipschitz constant for a gradient descent layer is larger. Practically, we find that only a few (\eg $4$ to $8$) iterations are required. For conjugate gradient and proximal layers, the inversion is accurate up to numerical precision, but requires the iterative forward process of the layer to be computed accurately for our method to also be.

In Sec.~\ref{ssec:cs_example} a modified soft thresholding function is used in place of the proximal operator for the $\ell_1$. While results (Fig.~\ref{fig:cs_example}) suggest the effect of our change is negligible, the performance of the reconstruction could be reduced to allow for the invertibility of the operation (Sec.~\ref{ssec:inverse_prox}) and use of our method. Depending on the slope added the soft thresholding function, the performance and invertibility are traded off. When the slope is very small (on the order of machine epsilon), the performance of the reconstruction will behave similar to the ordinary function, however, it will be less invertible due to floating point quantization. When the slope is larger, the reconstruction performance could be reduced because the operator does not well model the proximal function of the reconstruction objective, but will be more linear, thus be less affected by quantization and more invertible.

Acceleration layers to improve convergence of image reconstruction are commonly used in variants of PGD (termed FISTA~\cite{beck2009fast}) and can be incorporated into our framework. Typically, such layers linearly combine the output of the current and previous layers, so inherently, the acceleration layer cannot be inverted from only the current layer's output. However, with the storage of additional information (this layer's output and the previous layer's output) it is possible to invert an acceleration layer by computing the inverse of a $2\times2$ matrix.

Finally, a limitation of our method is when each smaller auto-differentiation graph (discussed in Sec.~\ref{sec:methods}) is still too large to fit in memory. In this situation more context-specific solutions (\eg coil compression for multi-channel MRI, using a smaller FoV for FPM) or more efficient implementation of the system's fundamental operations are required. Another method to reduce the memory per layer is to write custom operations for backpropagation rather than rely on auto-differentiation.

\section{Conclusion}
\label{sec:conclusion}

Memory-efficient learning with physics-based networks is a practical tool for large-scale computational imaging problems. Using the concept of reversibility, we implemented reverse-mode differentiation with favorable storage and computational complexities. We demonstrated our method on several representative large-scale applications: multi-channel MRI and super-resolution optical microscopy, and expect other computational imaging systems to fall within our framework.

\section*{Funding Information}

This work was supported by STROBE: A National Science Foundation Science \& Technology Center under Grant No. DMR 1548924, by National Science Foundation under award No. 1755326, and by the Gordon and Betty Moore Foundation’s Data-Driven Discovery Initiative through Grant GBMF4562 to Laura Waller (UC Berkeley). Laura Waller is a
Chan Zuckerberg Biohub investigator. Michael R. Kellman is
additionally supported by the National Science Foundation’s
Graduate Research Fellowship under Grant No. DGE 1106400.
Emrah Bostan's research is supported by the Swiss National
Science Foundation (SNSF) under grant P2ELP2 172278. Michael Lustig's research is supported by GE Health care and by National Institutes of Health under award No. R01EB026136, R01HL136965, and R01EB009690.

\bibliography{bibtex}
\bibliographystyle{IEEEtran}

% % use section* for acknowledgment
% \section*{Acknowledgment}


\end{document}